\pdfoutput=1

\documentclass[11pt]{article}

\usepackage[final]{acl}
\usepackage{times}
\usepackage{latexsym}
\usepackage{subcaption}
\usepackage[T1]{fontenc}

\usepackage[utf8]{inputenc}

\usepackage{microtype}

\usepackage{inconsolata}

\usepackage{graphicx}
\usepackage{amsmath}
\usepackage{multirow}

\usepackage{xspace}
\newcommand{\sys}{SharePrefill\xspace}

\usepackage{bm}
\usepackage{newtxmath}
\usepackage{booktabs}
\usepackage{comment}
\usepackage{algorithm}
\usepackage{algorithmic}

\definecolor{comment_color}{RGB}{128,128,128}
\newcommand{\LineComment}[1]{\vspace*{0.5em}\small\textcolor{comment_color}{\textit{\# #1}}}

\usepackage{colortbl}      
\definecolor{purple}{RGB}{200, 160, 220} 

\usepackage{appendix} 

\title{Accelerating Prefilling for Long-Context LLMs via Sparse Pattern Sharing}

\author{
Dan Peng$^{1}$, Zhihui Fu$^{1}$, Zewen Ye$^{2}$, Zhuoran Song$^{3}$, Jun Wang$^{1}$\\
$^1$OPPO Research Institute,
$^2$Zhejiang University,
$^3$Shanghai Jiaotong University\\
\texttt{lepangdan@outlook.com}, \texttt{hzzhzzf@gmail.com}, 
\texttt{zewen-ye@outlook.com},
\\
\texttt{songzhuoran@sjtu.edu.cn},
\texttt{junwang.lu@gmail.com}
}

\begin{document}

\maketitle

\begin{abstract} 
Sparse attention methods exploit the inherent sparsity in attention to speed up the prefilling phase of long-context inference, mitigating the quadratic complexity of full attention computation.
While existing sparse attention methods rely on predefined patterns or inaccurate estimations to approximate attention behavior, they often fail to fully capture the true dynamics of attention, resulting in reduced efficiency and compromised accuracy.
Instead, we propose a highly accurate sparse attention mechanism that shares similar yet precise attention patterns across heads, enabling a more realistic capture of the dynamic behavior of attention.
Our approach is grounded in two key observations: (1) attention patterns demonstrate strong inter-head similarity, and (2) this similarity
remains remarkably consistent
across diverse inputs. 
By strategically sharing computed accurate patterns across attention heads, our method effectively captures actual patterns while requiring full attention computation for only a small
subset of heads.
Comprehensive evaluations demonstrate that our approach achieves superior or comparable speedup relative to state-of-the-art methods while delivering the best overall accuracy.
The code will be made available upon publication.

\end{abstract}

\section{Introduction}\label{sec:intro}

\begin{figure}[t]
\centering
\includegraphics[width=0.5\textwidth]{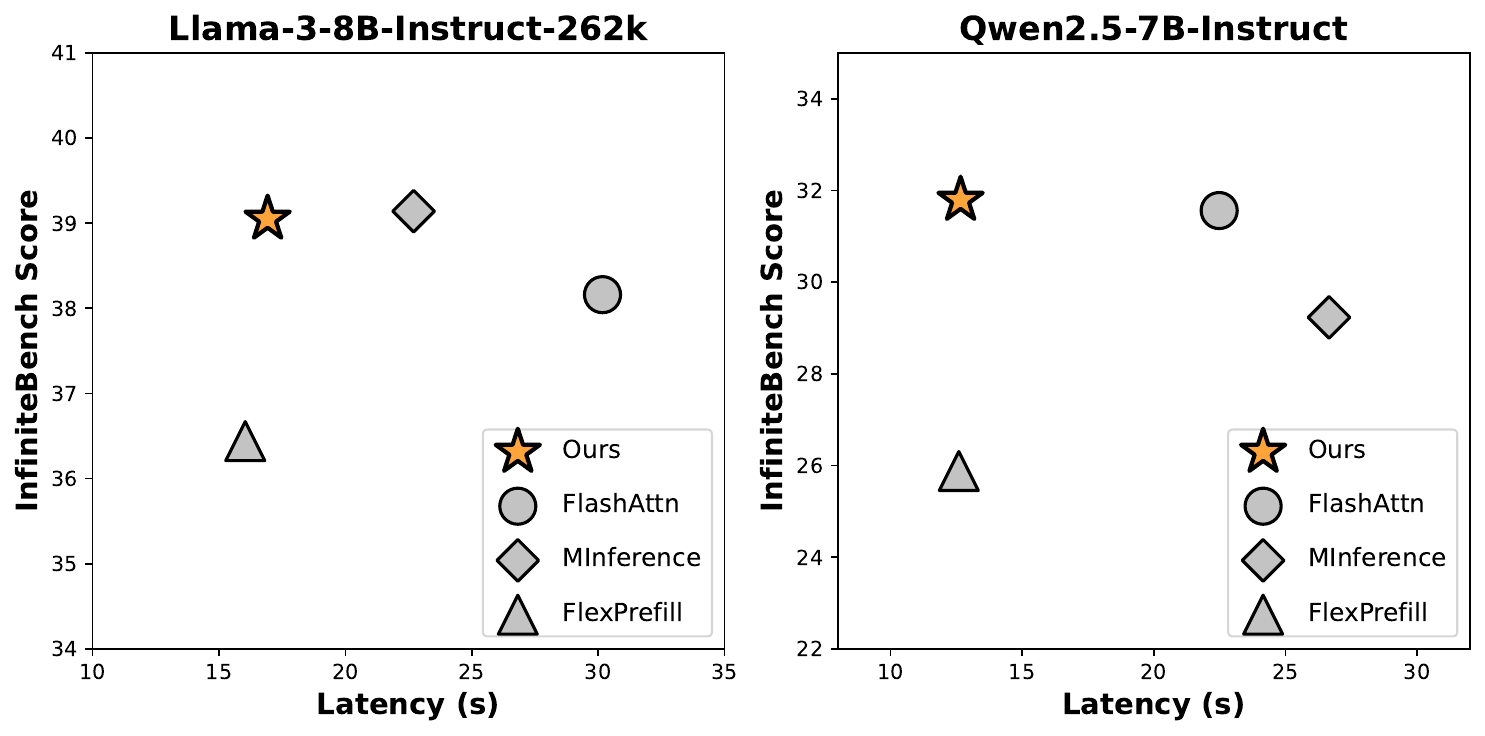}
\caption{
Comparison of our method with baselines across different models 
between latency (under 128K) and the average score on Infinitebench.
}
\label{fig:tradeoff}
\vspace{-10pt}
\end{figure}

Long-context inference is essential for real-world applications of large language models (LLMs).
Modern models like GPT-4.1 and Gemini 1.5~\cite{team2024gemini} now support contexts up to one million tokens, advancing multi-document QA \cite{wang2024knowledge}, code understanding \cite{bairi2024codeplan, ziftci2025migrating}, and multi-turn dialogue \cite{zhang2025multi}. Nonetheless, the prefilling phase of long-context inference remains time-consuming, as the vanilla attention mechanism entails quadratic computational complexity with respect to sequence length~\cite{fu2024challengesdeployinglongcontexttransformers}.

Sparse attention offers a promising solution by computing only significant attention scores, leveraging inherent sparsity in attention mechanisms. 
Many works have discovered various patterns with distinct characteristics and exploited them to perform sparse attention computations, 
such as the sink pattern in \cite{xiao2023efficient}, the A-shape, vertical-slash and block-sparse
patterns in MInference~\cite{jiang2024minference}.
However, these static patterns fail to generalize to varied inputs, as attention patterns inherently vary with different inputs as shown in~\autoref{fig:observe}. 
To cope with the fundamental requirements of dynamic patterns, MInference dynamically adjusts the vertical-slash index, and FlexPrefill~\cite{lai2025flexprefill} further adapts the vertical-slash sparsity ratio 
dynamically and uses pooled queries and pooled keys to estimate query-aware block-wise patterns.
However, we argue that pooling-based pattern estimation struggles to fully capture critical blocks due to the inaccuracies inherent in its approximations~(detailed in \autoref{sec:notenough}).
Alternatively, 
we discover two interesting phenomena. Firstly, the sparse pattern of many attention heads tends to be highly similar. 
More importantly, the similarity relationships among these heads remain largely consistent, even though the sparse patterns themselves vary significantly across different inputs, as shown in \autoref{fig:observe}. Consequently, 
we propose a 
highly accurate sparse attention mechanism that shares similar yet precise attention patterns across heads, mitigating reliance on predefined patterns and avoiding inaccurate pattern estimation.
By computing dense attention using only a subset of heads,
the prefilling is accelerated while preserving its high accuracy.

   
Our contributions are summarized as follows:
    \begin{itemize}
    \setlength{\itemsep}{0pt} 

    \item We empirically demonstrate two fundamental properties of sparse attention patterns: (1) similarity across attention heads and (2) similarity consistency across different inputs.

        \item We propose \sys, a novel 
        highly accuracy-preserving sparse attention method
        to accelerate the prefilling 
        phase by dynamically generating accurate sparse patterns and sharing them across heads.
        
        \item We conduct extensive experiments on several different models and tasks and show that our proposed method achieves superior or comparable speedup to state-of-the-art methods while achieving the best overall accuracy.
    \end{itemize}

\section{Related Work}\label{sec:related-work}

Existing sparse attention methods for accelerating  model inference can be categorized into two types: 
training-free
sparse attention and training-based sparse attention. 
The former relies on predefined sparse patterns or pattern estimation,
while the latter involves training sparse models to dynamically predict sparse patterns during inference.

\textbf{Training-free Sparse Attention} Several methods focus on predefined attention patterns, such as 
shifted sparse attention~\cite{chen2023longlora}, 
sink attention~\cite{xiao2023efficient} and  the A-shape, vertical-slash and block-sparse
patterns used in MInference~\cite{jiang2024minference}. However, these patterns, often derived from limited cases, lack the flexibility to effectively adapt to varying input demands. 
MInference introduced partially dynamic patterns, by adjusting vertical-slash indexes based on inputs. FlexPrefill~\cite{lai2025flexprefill} adapts sparsity ratios via cumulative thresholds and incorporates query-aware sparse patterns to enhance flexibility. 
However, query-aware sparse patterns rely on pooled query and key representations for pattern selection, which may cause information loss and lead to less accurate pattern estimation. Our method aligns with this line of work but further enhances pattern modeling by dynamically providing more precise sparse patterns through pattern sharing, thereby achieving better accuracy preservation.


\textbf{Training-based Sparse Attention} 
Training-based sparse attention methods introduce attention gates, train the gate-associated network, and automatically predict important sequence segments during inference.
In this series of works, approaches like MoBA~\cite{lu2025moba} and NSA~\cite{yuan2025native} continue training the entire model, while SeerAttention~\cite{gao2024seerattention} employs a linear layer as a learnable gate, training only the attention gate. Even though training-based sparse attention methods show promising acceleration while maintaining accuracy, the cost of resource-intensive and time-consuming training hinders their widespread practical applicability.


\section{Static Patterns and Pooling-based Pattern Estimation are Not Enough}~\label{sec:notenough}
Attention patterns are highly dynamic, showing substantial variation both across different heads and within the same head under different inputs, as shown in \autoref{fig:observe}. 
In particular, the staircase-like patterns in \textit{En.Dia} and the highly irregular patterns in \textit{Code.Debug} deviate significantly from previously proposed static patterns like the vertical-slash pattern~\cite{jiang2024minference}.
The highly dynamic nature intrinsic to attention mechanisms exposes the limitations of fixed-pattern approaches and underscores the need for adaptive, dynamic attention modeling techniques.


FlexPrefill~\cite{lai2025flexprefill} leverages pooled queries ($\boldsymbol{Q}$) and keys ($\boldsymbol{K}$) to estimate the average attention scores within each block for identifying critical regions, thus alleviating the reliance on predefined patterns. However, we highlight that this pooling-based method struggles to fully capture important blocks, and we identify that this challenge is rooted in two critical aspects.

\textbf{{Disregard for Token Alignment}:} 
The pooling operation disregards token-level position alignment within the query ($\boldsymbol{Q}$) and key ($\boldsymbol{K}$) segments, while attention mechanisms are inherently sensitive to token-level position alignment. This discrepancy leads to pooled results $pool(\boldsymbol{Q}) \cdot pool(\boldsymbol{K})$ that cannot accurately estimate the average of actual attention scores for the block. For example, consider two 1-dimensional $\boldsymbol{Q}$, $\boldsymbol{K}$ for 3 tokens: $\boldsymbol{Q}$=[0, 0, 1], $\boldsymbol{K}$=[0, 1, 0]. Due to ignoring position alignment, the $pool(\boldsymbol{Q}) \cdot pool(\boldsymbol{K})$ appears slightly significant($\frac{1}{9}$). However, all attention scores within the block are actually zero, leading to an \textit{overestimation} of the block's importance.

\textbf{{Smoothing of High-/Low- Values}:}
The pooling operation smooths out high and low values within $\boldsymbol{Q}$ and $\boldsymbol{K}$, which often contribute to high and low attention scores, resulting in inaccurate importance estimation.
For instance, $\boldsymbol{Q}$=[0, 0, 1], $\boldsymbol{K}$=[0, -1, 1]. During pooling, the 
the high-value and low-value elements in $\boldsymbol{Q}$ and $\boldsymbol{K}$ are diluted, 
resulting in $pool(\boldsymbol{Q}) \cdot pool(\boldsymbol{K}) = 0$, which is less than the actual average of attention scores $pool(\boldsymbol{Q} \cdot \boldsymbol{K}) = \frac{1}{9}$, 
leading to an \textit{underestimation} of the importance of the block.

\section{Observation: Dynamic Attention Heads Exhibit Similar Patterns and Static Similarity Relationships
}\label{sec:observe}

\begin{figure}[t] 
    \centering 
    \vspace{-5pt}
    \subcaptionbox{Visualization of attention patterns for different heads across various tasks. Each group of three columns corresponds to the heads within a specific task.}{\includegraphics[width = 0.49\textwidth]{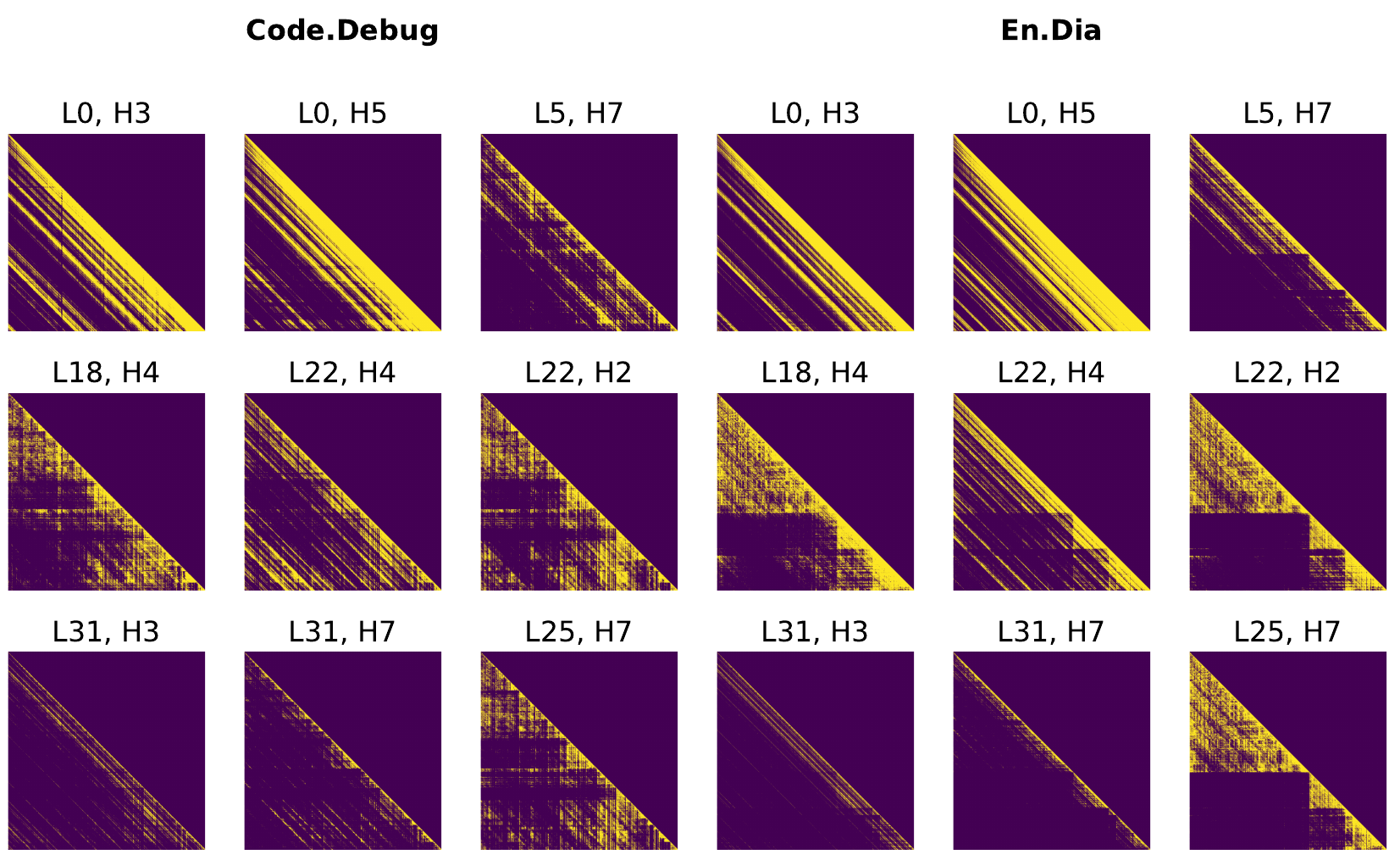}}
	\hfill
	\subcaptionbox{
    Similarity matrices show the pattern similarity between each head and other heads across different tasks.
    }
    {\includegraphics[width = 0.49\textwidth]{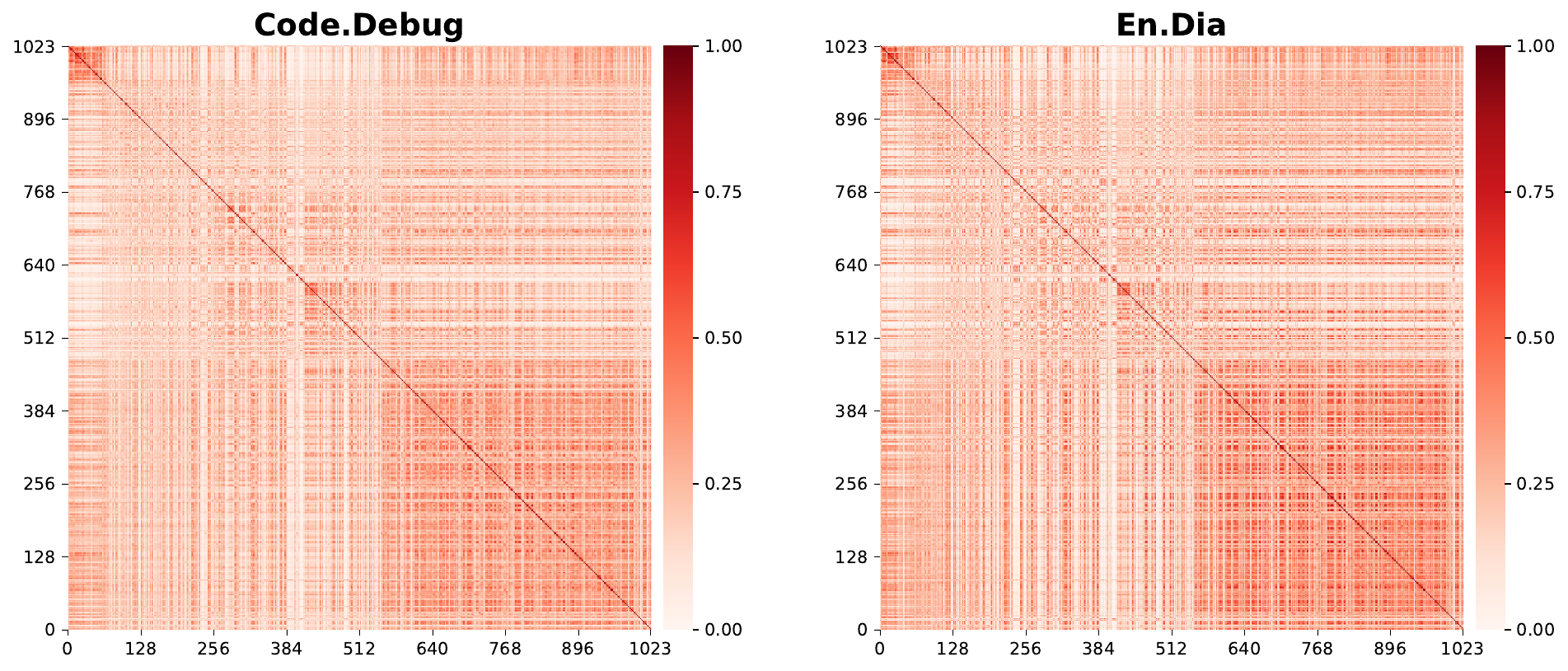}}
    \caption{Attention patterns of different heads and their similarity matrices across various tasks.}
    \label{fig:observe} 
    \vspace{-15pt}
\end{figure}




We present the foundational observations motivating our method: different heads exhibit high similarity, and the similarity remains highly consistent across varying inputs and tasks. Specifically, these observations distill into two key properties:


(1) \textit{\textbf{Inter-head} Pattern Similarity}:
    We observe many similar sparse patterns across attention heads, both within and between layers, as shown in~\autoref{fig:observe}(a). These patterns are derived from the Llama-3-8B-Instruct-262k model using samples from different tasks in InfiniteBench~\cite{zhang2024infinitebench}, with each group of three columns corresponding to heads from a specific task. For example, heads such as (L18, H4), (L22, H2), and (L25, H7) in the \textit{En.Dia} task exhibit highly consistent staircase-like patterns,  where L is the layer index and H is the head index.  Additionally, \autoref{fig:observe}(b) shows the statistical similarity matrix based on Jaccard similarity scores (\# intersection / \# union) between one head and all others. This measure avoids artificially high similarity values that could arise from the presence of many zeros in these sparse patterns. Notably, a large number of similarity scores exceed 0.5, indicating that each head has many similar counterparts among the others.
 
(2) \textit{\textbf{Cross-input} Similarity Consistency}:
    More importantly, the similarity among attention heads remains consistent regardless of the specific input or task, even though the pattern of a given attention head varies across different inputs and tasks. \autoref{fig:observe}(a) shows that (L18, H4), (L22, H2), and (L25, H7) are highly similar in \textit{Code.Debug}, consistent with their previously observed similarity in \textit{En.Dia} (Property 1), though their patterns differ between the two tasks. This highlights the consistent similarity among attention heads across inputs and tasks, suggesting that sparse patterns can potentially transfer to similar heads, regardless of context.

\section{Proposed Approach}\label{sec:approach}
\subsection{Problem Formulation}
Generally, our goal is 
to replace dense attention with sparse attention in attention layers to reduce computational costs during the pre-filling phase, while minimizing the output loss of each attention layer, thereby accelerating the pre-filling process while preserving accuracy.
This can be formulated as a multi-objective optimization problem:
\begin{equation}
\begin{split}
    \min_{\boldsymbol{M}} & \; |\mathcal{A}(\boldsymbol{Q}, \boldsymbol{K}, \boldsymbol{V}, \boldsymbol{M}) - \mathcal{A}(\boldsymbol{Q}, \boldsymbol{K}, \boldsymbol{V})|, \\
    \min_{\boldsymbol{M}} & \; |\boldsymbol{M}|
\end{split}
\end{equation}
where, \begin{align*}
    \mathcal{A}(\boldsymbol{Q}, \boldsymbol{K}, \boldsymbol{V}, 
    \boldsymbol{M}
    ) &= 
    \sigma
    \left(\frac{1}{\sqrt{d}} \boldsymbol{Q} \boldsymbol{K}^T - c(\boldsymbol{1}-\boldsymbol{M})\right) \boldsymbol{V}, \\
    \mathcal{A}(\boldsymbol{Q}, \boldsymbol{K}, \boldsymbol{V}) &= 
    \sigma
    \left(\frac{1}{\sqrt{d}} \boldsymbol{Q}\boldsymbol{K}^T\right) \boldsymbol{V}
\end{align*}

         
We define 
$\mathcal{A}(\boldsymbol{Q}, \boldsymbol{K}, \boldsymbol{V}, 
    \boldsymbol{M})$ as our sparse attention, where $\boldsymbol{M}$ is a binary mask indicating the sparse pattern applied in the sparse attention computation, where 1 means the block is computed, and 0 means it is skipped. The output is $\boldsymbol{O} = \mathcal{A}(\boldsymbol{Q}, \boldsymbol{K}, \boldsymbol{V}, 
    \boldsymbol{M}
    )$ and a sufficiently large constant $c$ ensures that attention score is approximately zero, whenever ${\boldsymbol{M}}_{ij} = 0$. Here, $\sigma$ denotes the softmax function.
The primary optimization objective is to identify a pattern $\boldsymbol{M}$ that minimizes the variance between the attention matrices of full attention
$ {\mathcal{A}}(\boldsymbol{Q}, \boldsymbol{K}, \boldsymbol{V}
    ) $
and our proposed sparse attention 
$ {\mathcal{A}}(\boldsymbol{Q}, \boldsymbol{K}, \boldsymbol{V}, 
    \boldsymbol{M}
    ) $
, while also reducing the computational time required for sparse attention computation 
and sparse pattern generation.

\subsection{Accelerating Prefilling via \sys}
Our main idea is to compute the full attention for a subset of heads, identify the actual sparse patterns, and share these patterns with other heads that are known to exhibit similar behavior. 
This approach enables the remaining heads to approximate the actual patterns without computing each one separately,
thus maintaining the model’s original accuracy while accelerating inference. 

\begin{figure*}[htbp] 
    \centering 
    \includegraphics[width=0.95\textwidth, height=0.45\textwidth]{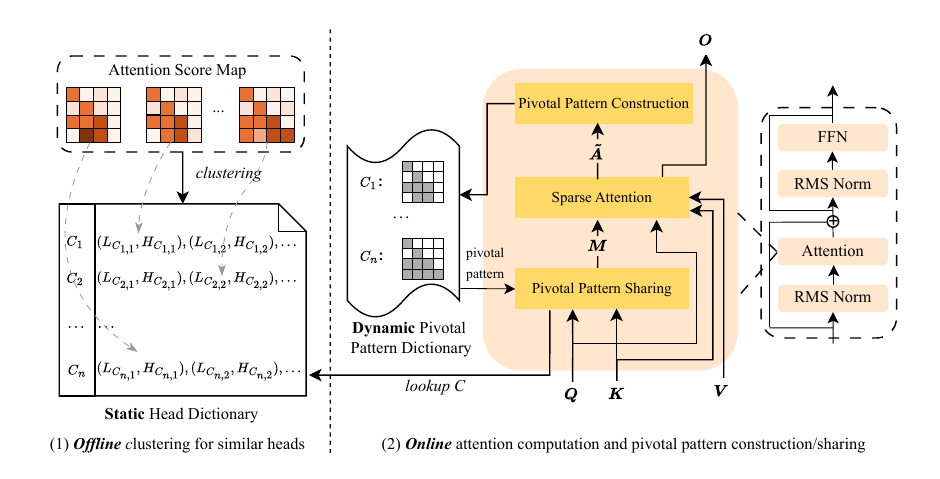} 
    \caption{
    Overview of proposed \sys. 
    Attention heads are clustered offline based on the similarity of their attention score maps to create a static head dictionary. During inference, each head retrieves its cluster index $C_i$. Pivotal Patterns are shared if available; otherwise, a dense pattern is assigned. The sparse attention output $\boldsymbol{O}$ is computed using $\boldsymbol{M}$, and $\boldsymbol{\tilde{A}}$ updates the dynamic pivotal pattern dictionary.
    }
    \label{fig:method_overview} 
\end{figure*}

As depicted in the overview of \sys in \autoref{fig:method_overview},
our approach involves two key components:~(1) Offline clustering to group heads based on the similarity of their attention score maps.
(2) Online inference, where pivotal attention is constructed dynamically and shared with other heads during the inference process. The overall
algorithm of \sys is detailed in Algorithm~\ref{alg:sparse_attn}.



\paragraph{Offline Clustering of Similar Heads}

We cluster attention heads across layers into distinct groups based on the similarity of their attention score maps, performed in offline mode.
The pre-computed clusters serve as the foundation for constructing and sharing sparse attention patterns within each cluster during inference.

Given the consistent
similarity between heads, we perform clustering on their attention score maps using a sample from the \textit{Retr.KV} task in InfiniteBench. We first obtain compressed low-dimensional representations of the attention scores by training an autoencoder network on these attention score maps (the network architecture is illustrated in \autoref{appendix:autoencoder}). Next, we normalize the representations and apply hierarchical clustering with a distance threshold to group similar heads into clusters, while isolating dissimilar heads as noise clusters. Notably, we only store the layer and head indices within clusters, rather than the sparse patterns themselves. The actual sparse patterns are dynamically generated during online inference, ensuring adaptability to varying inputs.

\paragraph{Dynamic Pattern Construction and Sharing during Inference}


During inference, we construct attention patterns dynamically and share adaptive yet accurate patterns among similar heads to facilitate sparse attention computation. This is achieved by computing the attention output for each layer while maintaining an evolving global pattern dictionary, which serves as the basis for sharing patterns across similar heads.

In general, the online inference process—based on dynamic pattern construction and sharing—comprises three key steps, corresponding to the three sub-algorithms outlined in Algorithm~\ref{alg:sparse_attn}.
The algorithm takes the query matrix $\boldsymbol{Q}$, key matrix $\boldsymbol{K}$, value matrix $\boldsymbol{V}$, 
similarity threshold $\tau$,  
sparsity threshold $\delta$
and cumulative attention threshold $\gamma$ as input. 
For simplicity, the illustration focuses on a single head rather than a layer. However, in practice, we perform sparse attention computation layer-by-layer. 


\textbf{Pivotal Pattern Sharing} ~(See Algorithm ~\ref{alg:share}):
Before performing sparse attention computation, we first query the global pivotal pattern dictionary to check if a pivotal pattern is available for reuse. If a pivotal pattern exists, it is shared with the corresponding head; otherwise, the head computes full attention using a dense pattern (i.e., a pattern with all ones).

\textbf{Sparse Attention Computation}~(See Algorithm ~\ref{alg:sparse_attn}): We then perform sparse attention computation to obtain the output for the current layer while simultaneously computing the block-wise average of QK values, denoted as $\boldsymbol{\tilde{A}}$, which captures the average QK scores within each block~(line 8 in overall Algorithm~\ref{alg:sparse_attn}). 
The sparse attention kernel is implemented in Triton~\cite{tillet2019triton}, following the block-wise strategy from FlashAttention 2~\cite{dao2024flashattention} and incorporating a block-wise sparse pattern to determine computation regions. Only blocks labeled as 1 in the sparse pattern are computed, while those labeled as 0 are skipped. During the computation for the final output, for each block where the pattern value is one, we compute
the average QK value; for blocks where the pattern value is 0, we assign the average QK value as $-\infty$. 




\textbf{Pivotal Pattern Construction}~(See Algorithm~\ref{alg:construct}):
Subsequently, we use the obtained block-wise average QK values to compute the block-wise average attention scores after applying softmax. These scores are then used to construct new pivotal patterns by applying a cumulative score threshold $\gamma$, which selects the minimal number of blocks required to cover the target cumulative attention score, as detailed in Algorithm~\ref{alg:construct}. The resulting patterns are then updated into the pivotal pattern dictionary.


To ensure safe dynamic pattern sharing, we verify similarity before sharing patterns to prevent incorrect sharing that could adversely impact accuracy. Specifically, we compute the Jensen-Shannon (JS)  distance between the block-wise average attention score of the last row block of the current head $\boldsymbol{\hat{a}}$ and the corresponding pivotal block-wise average attention score of the last row $\boldsymbol{\tilde{a}}$, which is also stored in the pivotal pattern dictionary~(line 6 in Algorithm~\ref{alg:determine}). This distance serves as a measure to predict the similarity between the current head and its corresponding pivotal head. If the JS distance is below a given similarity threshold $\tau$, we share the pivotal pattern with the current head. Otherwise, we fall back to a conservative vertical-slash pattern~(lines 7-11 in Algorithm~\ref{alg:determine}) using a cumulative threshold-based vertical-slash pattern search algorithm (outlined in Algorithm~\ref{alg:search_vs}), as proposed in FlexPrefill~\cite{lai2025flexprefill}. Additionally, noisy clusters, which include dissimilar patterns, also revert to the vertical-slash pattern.

To enhance efficiency, we exclude highly sparse heads from the pivotal pattern construction and sharing process, as we consider that computing full attention for these heads to derive pivotal patterns is not cost-effective in terms of acceleration. For highly sparse heads, we instead fall back to searching for a vertical-slash pattern for each head (see line 10 in Algorithm~\ref{alg:determine}), as the pattern often serves as a suitable approximation for highly sparse heads~\cite{jiang2024minference}. To identify these highly sparse heads, we compute the Jensen-Shannon (JS) distance between the block-wise average attention score of the last row block of the current head $\boldsymbol{\hat{a}}$ and a uniform distribution $\boldsymbol{u}$ (see line 6 in Algorithm~\ref{alg:determine}). We then compare this distance to a predefined sparsity threshold $\delta$. If the JS distance is not less than the threshold, we classify the head as a highly sparse head.
\begin{table*}[!htbp]
\centering
\begin{tabular}{p{1.0\textwidth}}
\toprule
\multicolumn{1}{c}{
\sys
} \\
\midrule
\midrule

\begin{tabular}{p{0.46\textwidth}p{0.52\textwidth}}
\label{alg:overall}


\begin{minipage}[t]{0.45\textwidth}
\vspace{-25pt}
\centering
\begin{algorithm}[H]
\captionsetup[algorithm]{singlelinecheck=off}
\caption{Sparse Attention}
\label{alg:sparse_attn}
\begin{algorithmic}[1]
  \STATE {\bfseries Input:} $\boldsymbol{Q}, \boldsymbol{K}, \boldsymbol{V} \in \mathbb{R}^{N \times d_{h}}$; 
  $\delta$, $\tau$, $\gamma$;
    $l$, $h$




    

    \LineComment{
    Decide the pattern type  based on $\boldsymbol{Q}$,
    $\boldsymbol{K}$, sparsity threshold $\delta$ and
    similarity threshold $\tau$
    }

     \STATE
     $pattern$
     $\gets$ 
     Determine
     Sparse Pattern ($\boldsymbol{Q}$,
    $\boldsymbol{K}$, $\delta$, 
    $\tau$)

    \LineComment{
    Decide the sparse pattern $\boldsymbol{M}$ based on \textit{pattern}, $\boldsymbol{Q}$, $\boldsymbol{K}$ and pattern threshold $\gamma$ 
    }
    \IF{$pattern$ == shared$\_$pivot
    }
   
    \STATE
    $\boldsymbol{M}$ $\gets$ 
    Share Pivotal Pattern($l$, $h$)
    \ELSIF{$pattern$ == vertical$\_$slash}
    \STATE
    $\boldsymbol{M}$ $\gets$ 
    Search Vertical Slash Pattern($\boldsymbol{Q}$,
    $\boldsymbol{K}$, 
    $\gamma$)
    \ENDIF


    \LineComment{Compute the  output $\boldsymbol{O}$  and block-averaged QK values $\boldsymbol{\tilde{A}}$ by applying sparse pattern $\boldsymbol{M}$
    } \\
    \STATE  $\boldsymbol{O}$,  $\boldsymbol{\tilde{A}}
    \gets$ 
    $\mathcal{A}$($\boldsymbol{Q}$, $\boldsymbol{K}$, $\boldsymbol{V}$, $\boldsymbol{M}$)
    \\
    

    \LineComment{
    Construct and update
    global dynamic pivotal patterns via the newest
    block-averaged QK values 
    $\boldsymbol{\tilde{A}}$ and pattern threshold $\gamma$ 
    }
    \STATE 

    Construct Pivotal Pattern($\boldsymbol{\tilde{A}}$, $\gamma$, $l$, $h$)
    
return $\boldsymbol{O}$
\end{algorithmic}
\end{algorithm}
\end{minipage}

&

\begin{minipage}[t]{0.48\textwidth}
\vspace{-16pt}
\centering
\begin{algorithm}[H]
\centering
\captionsetup[algorithm]{singlelinecheck=off}
\caption{
Construct Pivotal Pattern
}
\label{alg:construct}
\begin{algorithmic}
  \STATE {\bfseries Input:} $\boldsymbol{\tilde{A}}$; $\gamma$;  $l$, $h$

 \IF{ $\boldsymbol{\tilde{A}}$ is fully attention computed}
 \STATE
 \LineComment{
 Compute block-averaged attention score \(\boldsymbol{\tilde{A}}\) by applying softmax on block-averaged QK values 
 }
 \STATE
 $\boldsymbol{\tilde{A}}$ = softmax($\boldsymbol{\tilde{A}}$)

 \LineComment{
 Take the last row of $\boldsymbol{\tilde{A}}$ as pivotal representative
 }
 \STATE

 $\boldsymbol{\tilde{a}} \gets$  $\boldsymbol{\tilde{A}}_{[-1:]}$
 
    \LineComment{Flatten and normalize attention score map}
    \STATE
    $\boldsymbol{\tilde{A}}$ $\gets$  flatten($\boldsymbol{\tilde{A}} / \sum_{i,j} \boldsymbol{\tilde{A}}[I, j]$)

    \LineComment{
    Sort attention scores
    }
    
    \STATE
    $\boldsymbol{I_{a}}$ $\gets$ $argsort(\boldsymbol{\tilde{A}})$

   \LineComment{Obtain the minimum computational budget
making the sum of the scores exceeds $\gamma$
    }
    \STATE
    $K$ $\gets$ min $\{$k: $\sum_{i \in \boldsymbol{I_{a}}[1:k]} \boldsymbol{\tilde{A}}[i]$  $\geq$ $\gamma$ $\}$

    \LineComment{Select index set}
    
     \STATE
    $\boldsymbol{S} \gets$ $\boldsymbol{I_{a}}[1:K]$
    \\
    \LineComment{Convert index set $\boldsymbol{S}$ to mask pattern $\boldsymbol{M}$}
    \STATE
    $\boldsymbol{M} \gets$ index$\_$to$\_$mask($\boldsymbol{S}$)

    \LineComment{
    Lookup cluster index $c$ in \texttt{head\_dict}
    }
    \STATE 
     c $\gets$ lookup($l$, $h$; 
     \texttt{head\_dict})

    \LineComment{
    Update $\boldsymbol{M}$ and $\boldsymbol{\tilde{a}}$ into \texttt{pivotal\_pattern\_dict}
    }
     \STATE
     \texttt{pivotal\_pattern\_dict}.update(
     $\{$c: ($\boldsymbol{\tilde{a}}$, $\boldsymbol{M})\}$)

 \ENDIF

\end{algorithmic}
\end{algorithm}
\end{minipage}

\\

\begin{minipage}[t]{0.45\textwidth}
\vspace{-85pt}
\centering
\begin{algorithm}[H]
\captionsetup[algorithm]{singlelinecheck=off}
\caption{
Determine Sparse Pattern
}
\label{alg:determine}
\begin{algorithmic}[1]
  \STATE {\bfseries Input:}
  $\boldsymbol{Q}$,
$\boldsymbol{K}$; $\delta$, 
$\tau$

\LineComment{
Take a representative query subset
}

\STATE
    select $\boldsymbol{\hat{Q}}$  = $\boldsymbol{Q}_{[-block\_size:]}$
    
\LineComment{Compute estimated block-averaged
average attention $\boldsymbol{\hat{a}}$
and pivotal block-averaged attention~$\boldsymbol{\tilde{a}}$}

    \STATE 
    $\boldsymbol{\hat{a}}$ 
    $\gets$ softmax(pool($\boldsymbol{\hat{Q}}\boldsymbol{K}^{T} $)  $/\sqrt{d}$
    )
    \\




   \LineComment{Retrieve cluster index $c$ in \texttt{head\_dict}, 
   }

    \STATE 
     c $\gets$ lookup($l$, $h$; 
     \texttt{head\_dict}
     ) 
     \\
    \LineComment{Fetch the pivotal representative $\tilde{a}$}
     
     \STATE
     $\boldsymbol{\tilde{a}}$ $\gets$ lookup(c; 
     \texttt{pivotal\_pattern\_dict}
     )
     \\
     \LineComment{
     Compute sparsity  and similarity divergence
     }
      \STATE
     $d_{sparse}$ $\gets$  $\sqrt{JSD(
\boldsymbol{\hat{a}}
    ||
    \boldsymbol{u}
    )}$,  $d_{sim}$ $\gets$  $\sqrt{JSD(
\boldsymbol{\hat{a}}
    ||
    \boldsymbol{\tilde{a}}
    )}$

    \LineComment{Determine whether to use pattern sharing strategy
    }
    
    \IF{$d_{sparse}$ < $\delta$ and $d_{sim}$ < $\tau$}
    \STATE
    $pattern \gets$ shared$\_$pivot 


    


   
    \ELSE
    
    \STATE
    $pattern$ $\gets$ vertical$\_$slash

    \ENDIF
    
return $pattern$

  
  


    

\end{algorithmic}
\end{algorithm}
\end{minipage}

&

\begin{minipage}[t]{0.45\textwidth}
\vspace{-20pt}
\centering
\begin{algorithm}[H]
\captionsetup[algorithm]{singlelinecheck=off}
\caption{Share Pivotal Pattern}
\label{alg:share}
\begin{algorithmic}
  \STATE {\bfseries Input:} 
   $l$, $h$

\LineComment{Retrieve cluster index $c$ in \texttt{head\_dict}}

     \STATE      
     c $\gets$ lookup($l$, $h$; 
      \texttt{head\_dict}
     )
    
     \LineComment{Fetch the pivotal sparse pattern from the dynamic \texttt{pivotal\_pattern\_dict} $\boldsymbol{M}$
}
     \STATE
     $\boldsymbol{M}$ $\gets$ lookup(c; 
     \texttt{pivotal\_pattern\_dict}
     )
     
     \IF{$\boldsymbol{M}$ not exist}
         \STATE
         \LineComment{Assign a dense pattern to the first head within the cluster $c$ for subsequent full attention computation
         }
         \STATE
         $\boldsymbol{M}$ $\gets$ $\boldsymbol{ones}$
    \ELSE
     \STATE
     \LineComment{Share existing pivotal pattern $\boldsymbol{M}$}
        
\ENDIF
\\     
return $\boldsymbol{M}$

\end{algorithmic}
\end{algorithm}
\end{minipage}

\end{tabular} \\
\bottomrule
\end{tabular}
\end{table*}


\begin{table*}[!htbp]
\centering
\scriptsize

\setlength{\tabcolsep}{1pt}
\begin{tabular}{c|ccccccccccc|c}
    \toprule
    \textbf{Models} & \textbf{Methods} & \textbf{En.Sum} & \textbf{En.QA} & \textbf{En.MC} & \textbf{En.Dia} & \textbf{Zh.QA} & \textbf{Code.Debug} & \textbf{Math.Find} & \textbf{Retr.PassKey} & \textbf{Retr.Number} & \textbf{Retr.KV} & \textbf{Avg} \\
    \midrule
  
    \multirow{5}[2]{*}{\textbf{Llama-3-8B-Instruct-262k}} & FlashAttn & 25.88  & 8.63  & 67.69  & 5.00  & 12.66  & 20.81  & 26.57  & 100.00  & 100.00  & 14.40  & 38.16  \\
          & FlexPrefill & {19.91}  & \textbf{12.60} & 57.21  & 5.50  & 11.63  & {22.84}  & 20.86  & 100.00  & 100.00  & 13.80  & 36.44  \\
          & MInference & \textbf{25.51} & 8.50  & \textbf{65.94} & \underline{8.00} & \textbf{12.14} & 22.08  & \textbf{32.86} & 100.00  & 100.00  & {16.40}  & \underline{39.14}  \\

          & \cellcolor[rgb]{ .784,  .627,  .863} Ours 
          & \cellcolor[rgb]{ .784,  .627,  .863}\underline{20.24} 
          & \cellcolor[rgb]{ .784,  .627,  .863}8.00  
          & \cellcolor[rgb]{ .784,  .627,  .863}63.32  
          & \cellcolor[rgb]{ .784,  .627,  .863}\textbf{11.84} 
          & \cellcolor[rgb]{ .784,  .627,  .863}11.94  
          & \cellcolor[rgb]{ .784,  .627,  .863}\underline{24.11} 
          & \cellcolor[rgb]{ .784,  .627,  .863}\underline{30.00}  
          & \cellcolor[rgb]{ .784,  .627,  .863}\textbf{100.00} 
          & \cellcolor[rgb]{ .784,  .627,  .863}\textbf{100.00} 
          & \cellcolor[rgb]{ .784,  .627,  .863}\underline{21.00} 
          & \cellcolor[rgb]{ .784,  .627,  .863}39.05 \\

          & \cellcolor[rgb]{ .784,  .627,  .863}Ours ($\delta = 1.01$)
          & \cellcolor[rgb]{ .784,  .627,  .863}19.35  
          & \cellcolor[rgb]{ .784,  .627,  .863}\underline{11.74} 
          & \cellcolor[rgb]{ .784,  .627,  .863}\underline{64.63} 
          & \cellcolor[rgb]{ .784,  .627,  .863}{5.50}  
          & \cellcolor[rgb]{ .784,  .627,  .863}\underline{11.96} 
          & \cellcolor[rgb]{ .784,  .627,  .863}\textbf{28.17} 
          & \cellcolor[rgb]{ .784,  .627,  .863}{29.14} 
          & \cellcolor[rgb]{ .784,  .627,  .863}\textbf{100.00} 
          & \cellcolor[rgb]{ .784,  .627,  .863}\textbf{100.00} 
          & \cellcolor[rgb]{ .784,  .627,  .863}\textbf{23.00} 
          & \cellcolor[rgb]{ .784,  .627,  .863}\textbf{39.35} \\
    \midrule
    \multirow{4}[2]{*}{\textbf{Qwen2.5-7B-Instruct}} & FlashAttn & 15.53  & 3.18  & 35.81  & 10.50  & 3.95  & 14.47  & 38.57  & 100.00  & 93.56  & 0.00  & 31.56  \\
          & FlexPrefill & 14.20  & \textbf{3.09} & 31.88  & 8.00  & 3.54  & \underline{15.99} & 9.43  & \underline{97.29} & 75.42  & 0.00  & 25.88  \\
          & MInference & \underline{14.83} & 2.86  & \underline{34.93} & \textbf{9.00} & \underline{3.81} & 14.97  & \underline{38.29} & 96.78  & \underline{76.78} & 0.00  & \underline{29.23} \\
          & \cellcolor[rgb]{ .784,  .627,  .863}Ours & \cellcolor[rgb]{ .784,  .627,  .863}\textbf{15.31} & \cellcolor[rgb]{ .784,  .627,  .863}\underline{2.88} & \cellcolor[rgb]{ .784,  .627,  .863}\textbf{38.43} & \cellcolor[rgb]{ .784,  .627,  .863}\underline{8.50} & \cellcolor[rgb]{ .784,  .627,  .863}\textbf{3.99} & \cellcolor[rgb]{ .784,  .627,  .863}\textbf{17.26} & \cellcolor[rgb]{ .784,  .627,  .863}\textbf{44.57} & \cellcolor[rgb]{ .784,  .627,  .863}\textbf{99.49} & \cellcolor[rgb]{ .784,  .627,  .863}\textbf{87.46} & \cellcolor[rgb]{ .784,  .627,  .863}\textbf{0.00} & \cellcolor[rgb]{ .784,  .627,  .863}\textbf{31.79} \\


    \bottomrule
\end{tabular}
\caption{Performance comparison of different methods on various models and tasks on InfiniteBench. The best and second-best results are highlighted in \textbf{bold} and \underline{underlined}, respectively.}
\label{tbl:perf_cmp_inf}
\end{table*}

\section{Experiments
}\label{sec:experiment}
\subsection{
Settings}\label{sec:settings}
This section outlines the models, datasets, baselines, and implementation details 
of our method in comparison with baseline methods. Additional information is provided in \autoref{appendix:exp_settings}.



\begin{table*}[!tbp]
  \scriptsize
  \centering
    \setlength{\tabcolsep}{1pt}
    \begin{tabular}{c|cccccccccc|c|c}
    \toprule
    \textbf{Methods} & \textbf{En.Sum} & \textbf{En.QA} & \textbf{En.MC} & \textbf{En.Dia} & \textbf{Zh.QA} & \textbf{Code.Debug} & \textbf{Math.Find} & \textbf{Retr.PassKey} & \textbf{Retr.Number} & \textbf{Retr.KV} & \textbf{Avg} & \textbf{128K Latency (s)} \\
    \midrule
    Our w/o Sharing ($\tau$=0) & 19.68 & 11.86 & 63.76 & 9.00 & 11.65 & 23.86 & 25.14 & 22.00 & 100.00 & 22.00 & 38.70 & \underline{17.01}
    
    \\
    Our w/o Exclusion 
    ($\delta$=1.01)
    & 19.35  & 11.74  & 64.63  & 5.50  & 11.96  & 28.17  & 29.14  & 100.00  & 100.00  & 23.00  & \textbf{39.35}  & 20.02  \\
    Ours      & 20.24  & 8.00   & 63.32  & 11.84 & 11.94  & 24.11  & 30.00  & 100.00  & 100.00  & 21.00  & \underline{39.05}  & \textbf{16.92}  \\
    \bottomrule
    \end{tabular}%
  \caption{Performance of ablation methods evaluated using LLaMA-3-8B-Instruct-262K on InfiniteBench.}
  \label{tab:ablation}
\end{table*}%

\begin{figure}[t]
    \centering
    \begin{subfigure}[b]{0.4\textwidth}\includegraphics[width=\textwidth]{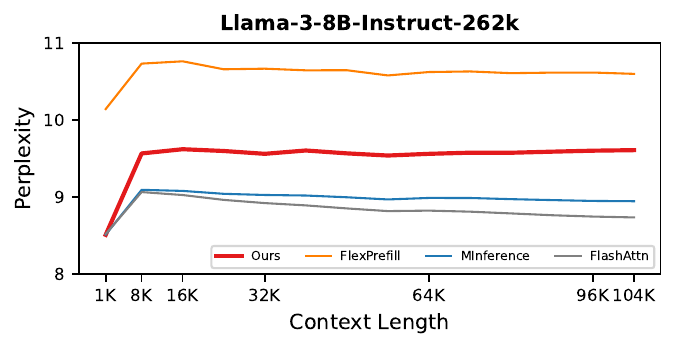}
    \end{subfigure}
    \begin{subfigure}[b]{0.4\textwidth} \includegraphics[width=\textwidth]{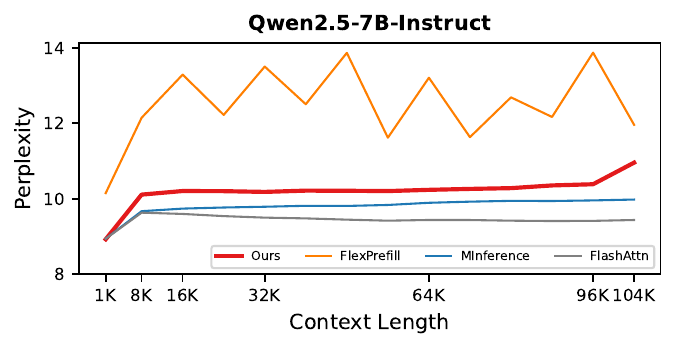}
    \end{subfigure}
    \caption{Perplexity results on PG-19~\cite{rae2019compressive} using different models and methods. }
    \label{fig:ppl}
\end{figure}

\paragraph{Models, Datasets, and Baselines}
We employ two 
cutting-edge, renowned
long-context LLMs: 
(i)
Llama-3-8B-Instruct-262k~\cite{gradientlongcontextllama3},
(ii) 
Qwen2.5-7B-Instruct~\cite{qwen25}. 
The models are evaluated on InfiniteBench~\cite{zhang2024infinitebench}, a state-of-the-art public benchmark designed to assess comprehensive long-context understanding.
This benchmark comprises synthetic and realistic tasks across various domains, in both English and Chinese, with an average token count of 214K, allowing us to assess \sys's effectiveness across a wide range of long-context scenarios.
Additionally, we include the long-context language modeling task on the PG19 dataset~\cite{rae2019compressive} to evaluate the models' language modeling capability. For efficiency evaluation, we conduct latency benchmarks using the length-adjustable prompts provided in MInference~\cite{jiang2024minference}.
We compare our method with two state-of-the-art sparse attention methods~\cite{jiang2024minference, lai2025flexprefill} and the efficient full attention FlashAttention 2~\cite{dao2024flashattention} to underscore its effectiveness and efficiency in long-context tasks.

\paragraph{Implementation Details}
All our experiments were conducted on a single NVIDIA A100 GPU with 80GB of memory.
For baseline implementations, we use the official FlashAttention 2 package~\ref{url:flash}, and adopt the official MInference repository~\ref{url:minf} for both MInference and FlexPrefill, which includes the FlexPrefill implementation.
For MInference, we employ the default vertical-slash pattern configuration available in its code repository. For FlexPrefill, we use the default parameters with the sparse pattern threshold \(\tau = 0.1\) and the cumulative pattern threshold \(\gamma = 0.9\) consistently for all models. For a fair comparison, we also set the cumulative pattern threshold \(\gamma = 0.9\) in our method; The similarity threshold $\tau$ is set to 0.2 and the sparsity threshold $\delta$ to 0.3, unless otherwise specified. Additionally, all the baseline methods employ sparse computation during prefilling and transition to dense computation during the decoding phase. 


\begin{figure}[t]
    \centering
    \begin{subfigure}[b]{0.46\textwidth}
    \includegraphics[width=\textwidth]{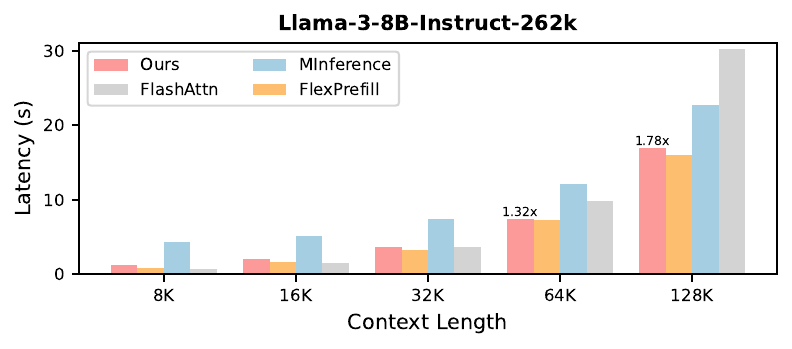}
    \end{subfigure}
    \begin{subfigure}[b]{0.46\textwidth} \includegraphics[width=\textwidth]{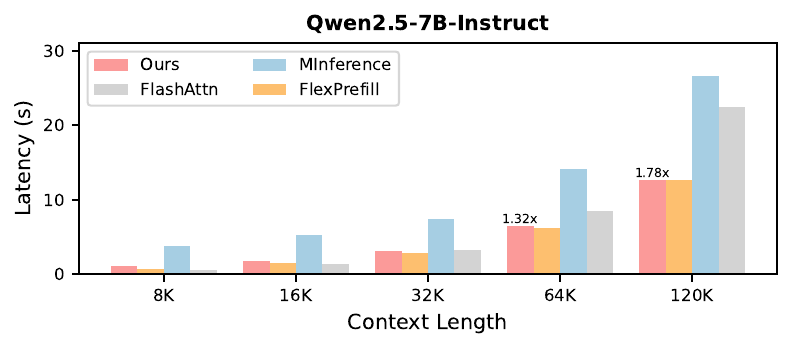}

    \end{subfigure}
    \caption{Latency comparison of different approaches across various context lengths using different models. }
   \label{fig:latency}
\end{figure}


\subsection{
Main Results
}
\label{sec:experiment-main_results}
We compare our method with baselines and present the main results on the aforementioned benchmarks and models.
The results demonstrate that our approach achieves superior or comparable speedup while delivering the overall best accuracy.

\paragraph{InfiniteBench}
\autoref{tbl:perf_cmp_inf} shows that our method preserves most of the model's performance, achieving overall best accuracy maintenance.
While our method with default parameters shows slightly lower accuracy compared to MInference, it achieves significantly lower latency, as shown in~\autoref{fig:latency} and~\autoref{fig:tradeoff}. However, our method outperforms MInference in both accuracy and efficiency by sharing all similar attention heads, including highly sparse ones.

\paragraph{Language Modeling}
We evaluate our method against baselines on the language modeling task based on the PG-19
dataset~\cite{rae2019compressive}. As shown in~\autoref{fig:ppl}, 
the perplexity of our method closely approaches the performance of MInference and FlashAttention 2, with the gap between them being within about 1.0. Moreover, the perplexity score of our method is significantly lower than that of FlexPrefill, with reductions of approximately 1.0$\sim$4.0 in Qwen2.5-7B-Instruct and over 1.0 in Llama-3-8B-Instruct-262k.
These results demonstrate the strong language modeling capabilities of our approach. 


\paragraph{Performance vs. Latency}
\autoref{fig:latency} 
shows the latency across different context windows under 
Llama-3-8B-Instruct-262k and Qwen2.5-7B-Instruct 
on a single A100. 
The results demonstrate that our method 
achieves better or comparable speedup compared to the baselines.
Moreover, \autoref{fig:tradeoff} compares our method with baselines 
under different models 
in terms of model performance on InfiniteBench and average latency under 128K on the latency benchmark. 
The results indicate that our method achieves a favorable tradeoff 
between accuracy preservation and inference speedup.







\begin{figure}[t]
    \centering
    \includegraphics[width=0.5\textwidth]{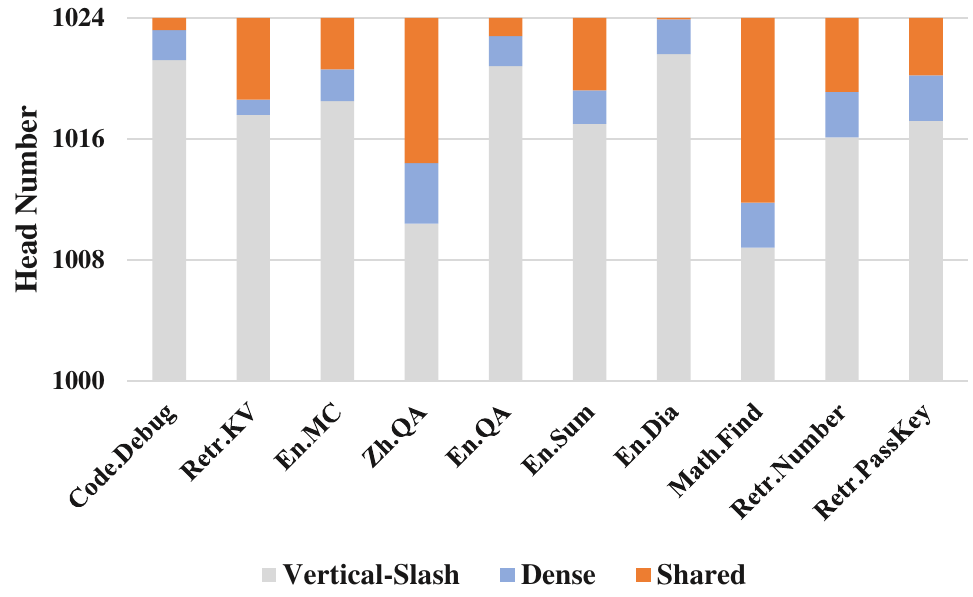}
    \caption{Distribution of three sparse attention head patterns in LLaMA-3-8B-Instruct-262K.}
   \label{fig:ptn_dis}
    \vspace{-5pt}
\end{figure}

\section{Ablation Study}\label{sec:ablation}

\paragraph{Analysis of Different Components}\label{para:ablation}
To evaluate the contributions of different components in \sys, we introduce two variants for the ablation study:
(1) Ours w/o sharing, which uses only the vertical-slash pattern without pivotal pattern sharing mechanism, corresponding to a similarity threshold $\tau$ = 0; (2) Ours w/o exclusion, where removing highly sparse heads strategy and all similar heads participate in the pattern sharing mechanism, corresponding to a sparsity threshold $\delta$ = 1.01 (selected to account for boundary conditions, ensuring that patterns with $\delta$ =1 meet the sharing-consideration criterion in line 7 of Algorithm~\ref{alg:determine}).
\autoref{tab:ablation} presents the ablation results on Llama-3-8B-Instruct-262k. It first demonstrates that removing the pattern sharing mechanism leads to performance degradation, confirming the necessity of our pattern sharing strategy in preserving accuracy. Additionally, removing the strategy of excluding highly sparse heads—where all similar heads, including highly sparse ones, are allowed to share patterns—results in reduced speedup but improved performance. This demonstrates that the strategy of excluding highly sparse heads enhances efficiency while potentially degrading the model's accuracy maintenance potential. The observed accuracy improvement when removing the exclusion strategy can be attributed to more similar heads participating in pattern sharing, rather than being forced into predefined vertical-slash patterns. This further validates the effectiveness of our pattern sharing mechanism in maintaining accuracy.
\paragraph{Pattern Distribution}
\autoref{fig:ptn_dis} shows the distribution of dense, shared, and conservative vertical-slash patterns used in Llama-3-8B-Instruct-262k. The majority of attention heads adopt the vertical-slash pattern, while only a small number require the full-attention dense pattern—typically just 1 to 4 heads in total. Although the number of shared patterns is limited, they play a significant role in maintaining model accuracy as shown in~\autoref{tab:ablation}.

\section{Conclusion and Future Work}\label{sec:conclusion}
In this paper, we observe that attention heads exhibit similarity, and this similarity remains consistent across different inputs.
Built on these observations, we propose a novel sparse attention method that highly preserves accuracy while accelerating the prefilling phase. Our method achieves this by dynamically generating accurate patterns and sharing them with other similar heads, thereby capturing more realistic attention dynamics.
We conduct extensive experiments across several models and tasks, demonstrating that our proposed method achieves superior or comparable speedups to state-of-the-art approaches while delivering the highest accuracy maintenance.
The principle of similarity between heads and the proposed pattern-sharing mechanism holds the potential for accelerating the decoding phase and extending to multi-modular systems, which will be explored in future work.

\newpage
\section*{Limitations}\label{sec:limit}
Although we provide observational and statistical evidence on the similarity properties between attention heads, the underlying explanation for the highly consistent similarity relationships among heads across different inputs remains unclear. This open question requires further investigation.
Additionally, while our approach demonstrates effectiveness in LLM prefilling on single devices, its scalability to larger-scale scenarios requires further study. Future work will focus on evaluating and further enhancing the scalability of the proposed approach. This includes exploring efficient pattern-sharing mechanisms in scaled scenarios, such as allowing each device to maintain a local partial pivotal pattern dictionary or enabling a global dictionary to be shared across devices through inter-device communication.
\bibliography{custom}

\begin{appendices}

\appendix
\section{More Details on Experimental Settings}
\label{appendix:exp_settings}


\subsection{Models}
We employ two state-of-the-art long-context language models: Llama-3-8B-Instruct-262k~\footnote{\url{https://huggingface.co/gradientai/Llama-3-8B-Instruct-Gradient-262k}\label{url:llama}}(released under the Meta Llama License) 
and Qwen2.5-7B-Instruct~\footnote{\url{https://huggingface.co/Qwen/Qwen2.5-7B-Instruct}\label{url:qwen}} (released under the Apache 2.0 License). 
These models were selected due to their strong capabilities in handling long-context understanding tasks, with Llama-3-8B-Instruct-262k supporting contexts of up to 262K tokens and Qwen2.5-7B-Instruct supporting contexts of up to 128K tokens. Both models support multiple languages, primarily English, with Qwen2.5-7B-Instruct also demonstrating excellent performance in Chinese. 
Additionally, Qwen2.5-7B-Instruct supports up to 128K tokens and demonstrates excellent multilingual performance, with particular strength in Chinese.
For further details, refer to the model repositories, as listed in~\ref{url:llama} and~\ref{url:qwen}

\subsection{Datasets $\&$ Benchmarks}
\begin{itemize}\setlength\itemsep{0.05em}
    \item \textbf{InfiniteBench}    InfiniteBench~\footnote{\url{https://huggingface.co/datasets/xinrongzhang2022/InfiniteBench}\label{url:infinitebench}}~\cite{zhang2024infinitebench} is publicly released under the Apache-2.0 License. It is a state-of-the-art benchmark designed to evaluate long-context language models with context lengths exceeding 100K tokens. The benchmark consists of 12 unique tasks, each carefully crafted to assess different aspects of language processing and comprehension in extended contexts. These tasks encompass a mix of real-world scenarios and synthetic constructs, including novels, dialogues, code, and math, ensuring a comprehensive evaluation of model capabilities.
    In our experiments, we compare our method's long-context performance against baselines across 10 tasks, using all available samples. Consistent with MInference and FlexPrefill, we excluded \textit{Code.Run} and \textit{Math.Calc} because they are highly challenging, with full-attention models often scoring near 0.
    
   \item  \textbf{PG-19 Language Modeling Benchmark
   }~\cite{rae2019compressive} proposed a long-context language modeling benchmark~\footnote{\url{https://github.com/google-deepmind/pg19}} 
   that evaluates perplexity on the PG-19 dataset and whose repository is released under the Apache 2.0 license. Perplexity quantifies how well a model predicts the next token in a sequence and is commonly used to assess the language modeling performance of long-context LLMs on extended texts. PG-19 contains books with lengths of up to 500K tokens, making it well-suited for long-context evaluation. 
   To assess language modeling performance across different context lengths, we conduct experiments using 100 randomly selected samples from the PG-19 dataset, truncating them to various lengths ranging from 1K to 104K tokens. We then report the average perplexity based on these truncated samples. Due to high memory usage during perplexity computation, we evaluate contexts with lengths of up to 104K tokens.
   \item \textbf{Latency Benchmark} 
   We follow the latency benchmarks provided in MInference~\footnote{\url{https://github.com/microsoft/MInference.git}\label{url:minf}}~\cite{jiang2024minference}, which is released under the MIT License. The prompts, sourced from the Chain-of-Thought Hub~\cite{fu2023chain} (also released under the MIT License), were trimmed to varying token lengths to measure the prefilling stage latency. To ensure reliable measurements, we conduct ten repeated experiments after a warm-up phase and report the average latency.
\end{itemize}
\subsection{Baselines}
\begin{enumerate}
    \item \textbf{FlashAttention 2}~\footnote{\url{https://pypi.org/project/flash-attn}\label{url:flash}}~\cite{dao2024flashattention}: 
    Flash Attention 2 is an I/O-aware exact attention algorithm designed to improve the efficiency of dense attention computation. It leverages tiling techniques to minimize the number of memory read and write operations between GPU high-bandwidth memory (HBM) and on-chip SRAM, thereby significantly reducing memory overhead and improving computational throughput. As an optimized implementation of dense attention, Flash Attention 2 enables faster and more scalable transformer inference and training, especially in long-sequence scenarios.
    \item \textbf{MInference}~\cite{jiang2024minference}: 
    MInference is a state-of-the-art sparse attention mechanism that exploits the static patterns observed in the attention mechanisms of LLMs, aiming to accelerate the prefilling phase for long-context inputs. It first determines offline which sparse pattern each attention head belongs to. During inference, it approximates the sparse indices online and dynamically computes attention using optimized custom kernels. This design enables significant speedup while maintaining strong accuracy.
    \item \textbf{FlexPrefill}~\cite{lai2025flexprefill}:
    FlexPrefill is another state-of-the-art sparse attention mechanism that enhances flexibility by incorporating cumulative-attention-based index selection and query-aware sparse patterns, enabling more adaptive sparse attention during the prefilling phase of LLM inference.
\end{enumerate}

\subsection{Implementation Details}
For offline clustering, we train an autoencoder on the attention score map with a latent dimension of 64. The model is trained for 1000 epochs with early stopping and a learning rate of 1e-3. We then apply the hierarchy clustering method fcluster from scipy~\footnote{\url{ https://scipy.org}} package on the normalized compressed representation using a distance threshold of 10, assigning clusters with fewer than 5 samples to a noise cluster.

\section{Detailed Algorithms}\label{appendix:alg}
\begin{algorithm}[!ht]
\centering
\captionsetup[algorithm]{singlelinecheck=off}
\caption{
Search Vertical Slash Pattern
}
\label{alg:search_vs}
\begin{algorithmic}
  \STATE {\bfseries Input:} 
  $\boldsymbol{Q}$, $\boldsymbol{K}$;
  $\gamma$
    \STATE 
    \LineComment{
    Compute a subset of the full attention map
    }

    \STATE 
    $\boldsymbol{\hat{A}}$ 
    $\gets$ softmax($\boldsymbol{\hat{Q}}\boldsymbol{K}^{T} $  $/\sqrt{d}$), where $\boldsymbol{\hat{Q}} \subset \boldsymbol{Q}$
    \\

     \LineComment{Sum and normalize attention scores along the
    vertical and slash directions}
    \STATE
    $\boldsymbol{a_{v}}$ 
    $\gets$ sum$\_$vertical($\boldsymbol{\hat{A}}) /$ $\sum_{i,j}\boldsymbol{\hat{A}}[I,j]$) 
    \\
    \STATE
    $\boldsymbol{a_{s}}$ 
    $\gets$ sum$\_$slash($\boldsymbol{\hat{A}}) /$ $\sum_{i,j}\boldsymbol{\hat{A}}[I,j]$) 
    \\

    \LineComment{
    Sort vertical and slash attention scores
    }
    \STATE
    $\boldsymbol{I_{v}}$ $\gets$ $argsort(\boldsymbol{a_{v}})$
    \STATE
    $\boldsymbol{I_{s}}$ $\gets$ $argsort(\boldsymbol{a_{s}})$

    \LineComment{
    Obtain the minimum computational budget
making the sum of the scores exceeds $\gamma$
    }
    \STATE
    $K_{v}$ $\gets$ min $\{$k: $\sum_{i \in \boldsymbol{I_{v}}[1:k]} \boldsymbol{a_{v}}[i]$  $\geq$ $\gamma$ $\}$

    \STATE
    $K_{s}$ $\gets$ min $\{$k: $\sum_{i \in \boldsymbol{I_{s}}[1:k]} \boldsymbol{a_{s}}[i]$  $\geq$ $\gamma$ $\}$

    \LineComment{Select vertical and slash index}

    \STATE
    $\boldsymbol{S_{v}}$ 
    $\gets$ $\boldsymbol{I_{v}}[1: K_{v}]$,  $\boldsymbol{S_{s}}$ 
    $\gets$ $\boldsymbol{I_{s}}[1: K_{s}]$
    \STATE
    $\boldsymbol{S} \gets$ $\boldsymbol{S_{v}} \cup \boldsymbol{S_{s}}$

    $\boldsymbol{M} \gets$ index$\_$to$\_$mask $\boldsymbol{S}$



    
return $\boldsymbol{M}$

\end{algorithmic}
\end{algorithm}



\section{Autoencoder Architecture}~\label{appendix:autoencoder}

Autoencoder architecture is shown in \autoref{tab:network_architecture}.
\begin{table*}[!h]
\centering
\begin{tabular}{lll}
\toprule
\multicolumn{3}{c}{\textbf{Encoder}} \\
\midrule
\textbf{Layer} & \textbf{Type} & \textbf{Parameters} \\
\midrule
Conv2d & Conv2D & out: 16, kernel: $3 \times 3$, padding: 1 \\
ReLU & Activation &  \\
MaxPool2d & Pooling & kernel: $4 \times 4$, stride: 4 \\
Conv2d & Conv2D & in: 16, out: 32, kernel: $3 \times 3$, padding: 1 \\
ReLU & Activation &  \\
MaxPool2d & Pooling & kernel: $4 \times 4$, stride: 4 \\
Flatten & Transformation & start\_dim=1, end\_dim=-1 \\
Linear & Fully Connected & in: 468512, out: 64 \\
\midrule
\multicolumn{3}{c}{\textbf{Decoder}} \\
\midrule
\textbf{Layer} & \textbf{Type} & \textbf{Parameters} \\
\midrule
Linear & Fully Connected & in: 64, out: 468512 \\
ReLU & Activation &  \\
Unflatten & Transformation & shape=(32, 121, 121) \\
ConvTranspose2d & Transposed Convolution & in: 32, out: 16, kernel: $4 \times 4$, stride: 2, padding: 1 \\
ReLU & Activation &  \\
ConvTranspose2d & Transposed Convolution & in: 16, out: 8, kernel: $4 \times 4$, stride: 2, padding: 1 \\
ReLU & Activation &  \\
ConvTranspose2d & Transposed Convolution & in: 8, out: 1, kernel: $4 \times 4$, stride: 4 \\
Sigmoid & Activation &  \\
\bottomrule
\end{tabular}
\caption{Network Architecture of the Autoencoder}
\label{tab:network_architecture}
\end{table*}

\end{appendices}
\end{document}